%% file: main.tex
\pdfoutput=1
\documentclass[conference]{IEEEtran}
\usepackage{booktabs} 

\usepackage[utf8]{inputenc} 
\usepackage[T1]{fontenc}    
\usepackage{hyperref}       
\usepackage{url}            
\usepackage{booktabs}       
\usepackage{amsfonts}       
\usepackage{nicefrac}       
\usepackage{microtype}      
\usepackage{lipsum}

\usepackage{graphicx}
\graphicspath{ {images/} }

\begin{document}
\twocolumn{
\centering
\title{Addressing the Fundamental Tension of PCGML with~Discriminative~Learning}

\author{\IEEEauthorblockN{Isaac Karth}
\IEEEauthorblockA{\textit{Dept. of Computational Media} \\
\textit{University of California, Santa Cruz}\\
1156 High Street\\
Santa Cruz, CA 95064\\
\texttt{ikarth@ucsc.edu}}
\and
\IEEEauthorblockN{Adam M. Smith}
\IEEEauthorblockA{\textit{Dept. of Computational Media} \\
\textit{University of California, Santa Cruz}\\
1156 High Street\\
Santa Cruz, CA 95064\\
\texttt{amsmith@ucsc.edu}}}

\maketitle
}

\begin{abstract}
Procedural content generation via machine learning (PCGML) is typically framed as the task of fitting a generative model to full-scale examples of a desired content distribution. This approach presents a fundamental tension: the more design effort expended to produce detailed training examples for shaping a generator, the lower the return on investment from applying PCGML in the first place. In response, we propose the use of discriminative models (which capture the validity of a design rather the distribution of the content) trained on positive and negative examples. Through a modest modification of WaveFunctionCollapse, a commercially-adopted PCG approach that we characterize as using elementary machine learning, we demonstrate a new mode of control for learning-based generators. We demonstrate how an artist might craft a focused set of additional positive and negative examples by critique of the generator's previous outputs. This interaction mode bridges PCGML with mixed-initiative design assistance tools by working with a machine to define a space of valid designs rather than just one new design.
\end{abstract}

\input{pcgml}
\bibliographystyle{IEEEtran}
\bibliography{wfc_pcgml}

\end{document}

%% file: pcgml.tex
\section{Introduction}

Procedural Content Generation via Machine Learning (PCGML) is the recent term for the strategy of controlling content generators using examples~\cite{DBLP:journals/corr/SummervilleSGHH17}. Existing PCGML approaches train their statistical models based on pre-existing artist-provided samples of the desired content. However, there is a fundamental tension here: machine learning often works better with more training data, but the effort to produce quality training data is frequently costly enough that the artists might be better off just making the content themselves.

Rather than attempting to train a generative statistical model (capturing the distribution of desired content), we focus on applying discriminative learning. In discriminative learning, the model learns to judge whether a candidate content artifact would be valid or desirable, but it does not learn how to generate candidates. Pairing a discriminative model with a pre-existing content generator, we realize example-driven generation that can be influenced by both positive and negative examples of valid design patterns. We examine this idea inside of an already-commercially-adopted example-based generation system, WaveFunctionCollapse (WFC)~\cite{Karth:2017:WCS:3102071.3110566}.\footnote{\url{https://github.com/mxgmn/WaveFunctionCollapse}} This approach begins to address the fundamental tension in PCGML while also opening connections to mixed-initiative design tools.

Mixed-initiative content generation tools~\cite{togelius2016chapter11} are designed around the idea of the artist having a conversation with the tool about one specific design across many alterations, with the goal of creating one high-quality design. We propose to adapt this \emph{conversational teaching model} for application in PCGML systems. While still having conversations with the tool about specific designs, the conversations are leveraged to talk about the general shape of the design space. The artist trains the generator to the point where it will be trusted to follow that style in the future, when the generator can be run non-interactively. Instead of an individual artifact, the goal is define a space of desirable artifacts from which the generator may sample.

It might seem most obvious to setup PCGML by using a generative statistical model. Such models explicitly capture the desired content distribution: $p(C)$. In the the proposed discriminative learning strategy, we intend only to learn whether a candidate design is considered valid or not: $p(V|C)$. Such models can often be fit from much less data than generative models, and they do not require the distribution of content $C$ in the samples to be representative of the target content distribution (easing the fundamental tension of PCGML). Indeed, many negative examples which demonstrate invalid designs will contain design details which should have zero likelihood. We assume the existence of a generator that can draw samples from a prior $p(C)$ filtered by the condition $p(V=true|C)$, effectively applying Bayes rule to draw samples from the conditional distribution: $P(C|V=true)$. WFC is one such generator that can be directly constrained to yield only designs that would be classified as valid.

This paper illuminates the implicit use of machine learning in WFC, explains how discriminative learning may be integrated, and presents a detailed worked example of the conversational teaching model. We refer to the primary user as an artist to emphasize the primarily visual interface. It should be understood that the process of creating the input image can involve both design skills and programming reasoning: the artist is specifying both an aesthetic goal and a complex system of constraints to achieve that goal. This is well within the scope of a technical artist's skill, and computer artists have a long history of leveraging complex computer science approaches in pursuit of aesthetic aims.

\section{Background}

In this section, we review WFC as an example-driven generator, characterize PCGML work to date as operating on only positive examples, and review the conversational interaction model used in mixed-initiative design tools.

\subsection{WaveFunctionCollapse}

WaveFunctionCollapse is a content generation algorithm devised by independent game developer Maxim Gumin. In contrast with generators presented in technical games research venues, WFC has seen surprisingly quick adoption within the technical artist community. Particularly notable is that WFC can be considered an instance of PCGML, as we illustrate in Sec.~\ref{sec:wfc_as_pcgml}.

The recently released Viking invasion game \emph{Bad North}~\cite{badNorth} uses Oscar St\r{a}lberg's WFC implementation for generating island maps.\footnote{As discussed in e.g. @OskSta: ``The generation algorithm is a spinoff of the Wave Function Collapse algorithm. It's quite content agnostic. I have a bunch of tweets about it if you scroll down my media tweets'' \url{https://twitter.com/OskSta/status/931247511053979648}} \emph{Caves of Qud}~\cite{cavesOfQud}, a roguelike that is currently in Early Access on Steam, uses WFC as one of its map generation techniques. The \emph{Caves of Qud} developers have closely followed the ongoing development of the algorithm, incorporating its recent improvements.\footnote{@unormal: ``Got Qud's new 20x faster WFC implementation down to about 50mb in-memory static overhead on init with 0 allocation for all future runs (unless the size of the gen output gets bigger, but Qud's doesn't), so no GC churn. (cc @ExUtumno )'' \url{https://twitter.com/unormal/status/984713110257852416}} In particular, Caves of Qud's implementation of the improved ``fast WFC'' enables it to use a higher $N$ for its $N\times N$ patterns, which potentially allows the developers to express more complex structures.\footnote{@unormal, 2:05 AM - 13 Apr 2018: ``20x faster also makes higher orders of N practical, which enables larger scale structures to pop out of wfc.'' \url{https://twitter.com/unormal/status/984719207156862976}}

WFC is an instance of content generation using constraint solving techniques~\cite{Karth:2017:WCS:3102071.3110566}. WFC analyzes an input image and expresses a weighted constraint satisfaction problem based on these local similarity properties. There are many alternative constraint solving systems that can be substituted for Gumin's original observe-and-propagate cycle, including our own declarative implementations with the answer-set solver Clingo~\cite{DBLP:journals/corr/GebserKKS14} and the recent ``fast WFC'' implementation by Mathieu Fehr and Nathana\"{e}l Courant~\cite{fastWFC}.

In this paper, we seek to extend the ideas of WFC while keeping them compatible with the existing implementations. One of the more unique aspects of WFC is that it is an example-based generator that can generalize from a single, small example image. In Sec.~\ref{sec:worked_example} we show that while more than one example is needed to appropriately sculpt the design space, the additional examples can be even smaller than the original and can be created in response to generator behavior rather than collected in advance.

\subsection{PCGML}

Summerville et al.\ define Procedural Content Generation via Machine Learning (PCGML) as the ``generation of game content by models that have been trained on \emph{existing} game content [emphasis added]''~\cite{DBLP:journals/corr/SummervilleSGHH17}. In contrast with search-based and solver-based approaches which presume the user will provide an evaluation procedure or logical definition of appropriateness, PCGML uses a more artist-friendly framing that assumes concrete example artifacts as the primary inputs. PCGML techniques may well apply constructive, search, or solver-based techniques internally after interpreting training examples.

Machine learning needs training data, and one significant source of data for PCGML research is the Video Game Level Corpus (VGLC), which is a public dataset of game levels \cite{DBLP:journals/corr/SummervilleSMV16}. The VGLC was assembled to provide corpora for level generation research, similar to the assembled corpora in other fields such as Natural Language Processing. In contrast with datasets of game level appearance such as VGMaps,\footnote{\url{http://vgmaps.com/}} content in the VGLC is annotated at a level suitable for constructing new, playable level designs (not just pictures of level designs).

The VGLC provides a valuable set of data sourced from iconic levels for culturally-impactful games (e.g.\ \emph{Super Mario Bros} and \emph{The Legend of Zelda}). It has been used for PCG research using autoencoders \cite{jain2016autoencoders}, generative adversarial networks (GANs) \cite{volz2018evolving}, long short-term memories (LSTMs) \cite{DBLP:journals/corr/SummervilleM16}, multi-dimensional Markov chains \cite[Sec.~3.3.1]{snodgrass2018markov}, and automated game design learning \cite{Osborn:2017:AMN:3102071.3110576}.

Summerville et al.\ identify a ``recurring problem of small datasets'' \cite{DBLP:journals/corr/SummervilleSGHH17}: most data only applies to a single game, and even with the efforts of the VGLC the amount of data available is small, particularly when compared to the more wildly successful machine learning projects. This is compounded by our desire to produce useful content for novel games (for which no pre-existing data is available). Hence the fundamental tension in PCGML: asking an artist (or a team of artists) to produce quality training data at machine-learning scale could be much less efficient than just having the artists make the required content themselves.

Compounding this problem, a study by Snodgrass et al.\ \cite{snodgrass2017studying} showed that the expressive volume of current PCGML systems did not expand much as the amount of training data increased. This suggests that the generative learning approach taken by these systems may not ever provide the required level of artist control. While this situation might be relieved by using higher-capacity models, the problem of the effort to produce the training data remains.

PCGML should be compared with other forms of example-based generation. Example-based generation long predates the recent deep learning approaches, particularly for texture synthesis. To take one early example, David Garber's 1981 dissertation proposed a two-dimensional, Markov chain, pixel-by-pixel texture synthesis approach~\cite{Garber:1981:CMT:910609}. Separate from Garber,\footnote{Efros and Leung later discovered Garber's previous work~\cite{Efros:2001:IQT:383259.383296}.} Alexei Efros and Thomas Leung contributed a two-dimensional, Markov-chain inspired synthesis method: as the synthesized image is placed pixel-by-pixel, the algorithm samples from similar local windows in the sample image and randomly chooses one, using the window's center pixel as the new value~\cite{efros1999texture}. Although WFC-inventor Gumin experimented with continuous color-space techniques descending from these traditions, his use of discrete texture synthesis in WFC is directly inspired by Paul Merrell's discrete model synthesis and Paul Harrison's declarative texture synthesis~\cite{Gumin2016readme}. Harrison's declarative texture synthesis exchanges the step-by-step procedure used in earlier texture synthesis methods for a declarative texture synthesis approach, patterned after declarative programming languages~\cite[Chap.~7]{harrison2006image}. Merrell's discrete 3D geometric model synthesis uses a catalog of possible assignments and expresses the synthesis as a constraint satisfaction problem~\cite{5557871}.

Unlike later PCGML work, these example-based generation approaches only need a small number of examples, often just one. However, each of these approaches use only positive examples without any negative examples.

\subsection{Mixed-Initiative Design Tools}

Several mixed-initiative design tools have integrated PCG systems. Their interaction pattern can be generalized as an iterative cycle where the generator produces a design and the artist responds by making a choice that contradicts the generator's last output. When the details of a design are underconstrained, most mixed-initiative design tools will allow the artist to re-sample alternative completions.

Tanagra is a platformer level design tool that uses reactive planning and constraint solving to ensure playability while providing rapid feedback to facilitate artist iteration~\cite{smith2011tanagra}. Additionally, Tanagra maintains a higher-order understanding of the beats that shape the level's pacing, allowing the artist to directly specify the pacing and see the indirect result on the shape of the level being built.

The SketchaWorld modeling tool introduces a declarative ``procedural sketching'' approach ``in order to enable designers of virtual worlds to concentrate on stating what they want to create, instead of describing how they should model it''~\cite{Smelik:2011:SMC:1961704.1961746}. The artist using SketchaWorld focuses on sketching high-level constructs with instant feedback about the effect the changes have on the virtual world being constructed. At the end of interaction, just one highly-detailed world results.

Similarly, artists interact with Sentient Sketchbook via map sketches, with the generator's results evaluated by metrics such as playability. As part of the interactive conversation with the artist, it also presents evolved map suggestions to the user, generated via novelty search~\cite{6932873,liapis2013sentient}. Although novelty search could be used to generate variations on the artist's favored design, it is not assumed that all of these variations would be considered safe for use. Tools based on interactive evolution do not learn a reusable content validity function nor do they allow the artist to credit or blame a specific sub-structure of a content sample as the source of their fitness feedback. In Sec.~\ref{sec:discriminative_learning}, we demonstrate an interactive system that can do both of these.

As these examples demonstrate, mixed-initiative tools facilitate an interaction pattern where the artist sees a complete design produced by a generator and responds by providing feedback on it, which influences the next system-provided design. This two-way conversation enables the artist to make complex decisions about the desired outcome without requiring them to directly master the technical domain knowledge that drives the implementation of the generator.   

The above examples demonstrate the promising potential of design tools that embrace this mode of control. However, they tend to focus on using generation to assist in creating specific, individual artifacts rather than the PCGML approach of modeling a design space or a statistical distribution over the space of possible content.

Despite the natural relationship between artist-supplied content and the ability of machine learning techniques to reflect on that content and expand it, PCGML-style generators that learn during mixed-initiative interaction have yet to be explored.

\section{Characterizing WFC as PCGML}
\label{sec:wfc_as_pcgml}

\begin{figure*}
  \includegraphics[width=7in]{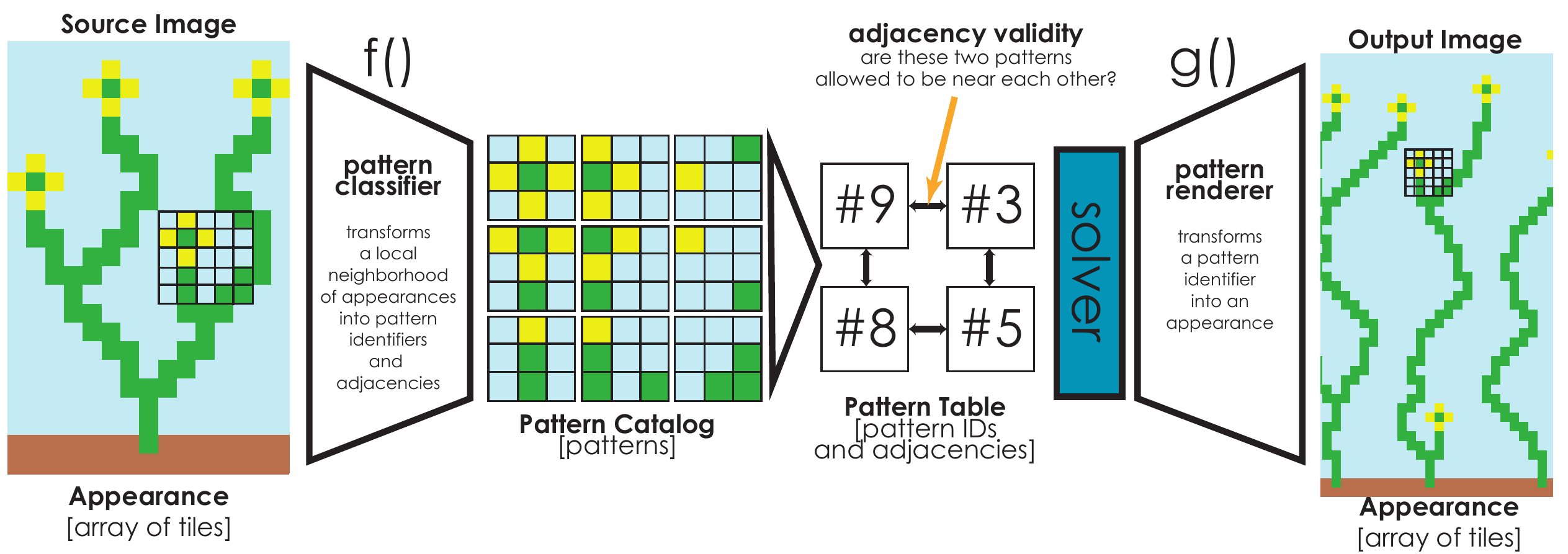}
\caption{\emph{Overview of WaveFunctionCollapse.} Gumin's OverlappingModel starts with a source image composed of colored tiles. Rather than operating directly on the tiles, the \emph{pattern classifier} transforms each local neighborhood into an identifier in a \emph{pattern catalog}, where each element is a unique $N\times N$ tile neighborhood. The \emph{adjacency validity} for each pair of patterns at each offset is also recorded (representing a Boolean function with two pattern identifier arguments). These patterns are used to define the constraints for the \emph{generator}, which uses the constraint solver to propagate pattern placements. Finally, the \emph{pattern renderer} is needed to transform the array of identifiers back into an appearance. This can be as simple as using the center tile of the $N\times N$ pattern. The italicized terms above are being introduced by this paper.}
\label{fig:stages}
\end{figure*}

Snodgrass describes WaveFunctionCollapse as ``an example of a machine learning-based PCG approach that does not require a deep understanding of how the algorithm functions in order to be used effectively''~\cite[Sec.~2.8]{snodgrass2018markov}. This section describes what and how WFC learns in a generalized vocabulary, which we introduce in this paper, that opens up space for exploring alternative learning strategies.

Gumin's original WaveFunctionCollapse algorithm proceeds in two phases. In the first, the single input image is analyzed to identify a vocabulary of local tile patterns and the possible adjacencies between those patterns. In the second phase, the results of the analysis are used as constraints in a generation process identifiable as constraint solving~\cite{Karth:2017:WCS:3102071.3110566}. In this section, we interpret the pattern analysis phase as an instance of machine learning. In particular, we show that this phase learns three functions. One classifies which pattern is present at each location (the pattern classifier function). One identifies if a pattern adjacency pairing is within the artist's preferred style (the adjacency validity function). The final function (the pattern renderer function) determines how a location in the output should be rendered given the the constraint-solving generator's choice of pattern placement assignments. Fig.~\ref{fig:stages} illustrates the relationship between these three functions.

\subsection{Pattern Classifier}

The original implementation of WFC has two models: a SimpleTileModel that uses explicit artist-specified adjacency relationships between tiles (expressed in an XML file) and an OverlappingModel that learns tile adjacencies from a source image. The OverlappingModel starts with a pattern identification step, which analyzes the input image for $N\times N$ patterns of tiles (where N is configurable, typically 2 or 3) which are the basic input for its constraint solving approach. The function that transforms a local patch of tiles into a pattern identifier is effectively learned by populating a lookup table. Even though this is a trivial method for constructing the classifier function, it is useful to see it as explicitly learning to identify where alternative learning methods might be used instead.

While the lookup table approach is effective in WFC, it implicitly disallows the use of any arrangement of tiles that was not seen in the source image (it would have no corresponding pattern identifier onto which it could map). In the generalized setting, we can imagine a deep convolutional neural network being used to map as-yet unseen tile configurations into the existing pattern catalog so long as they were perceptually similar enough.

Pattern classification need not be a strictly local operation. If we wanted to generate dungeon levels for a roguelike game, we may be particularly interested to note the distinction between treasure chests which are easily reachable by the player or not. A global analysis algorithm might assign different pattern identifiers to identically-appearing regions based on whether the region should be considered visible or hidden to the player. In platformer level generation, this could distinguish coins or other rewards placed on the player's default path (as a guide) or off the path (as an enticement to explore).

In the future, we imagine the contextual information used in the pattern classifier to come from many different sources. In the \emph{texture-by-numbers} application of image analogies~\cite{hertzmann2001image}, the artist hand-paints an additional input image to guide the interpretation of the source image and the generation of the target image in another example-driven image generator. In Snodgrass' hierarchical approach to tile-based map generation~\cite{snodgrass2015hierarchical}, a lower-resolution context map is generated automatically using clustering of tile patterns.

The goal of the pattern classifier is to assign, for every location in an input image, a pattern identifier number. Although the input to this function has an application specific datatype, we require that the output is always a bounded integer (e.g.\ from 1 to the number of patterns in the catalog). The job of the classifier is to collect pertinent details about what is happening in that location of the image and the surrounding context into a single value.

\subsection{Pattern Renderer}

Working inversely from the pattern classifier, the final step of WFC is to translate a grid of pattern selections back into a grid of tiles. The original implementation of WFC takes the straightforward approach of directly reusing the tile that is in the center of the stored $N\times N$ pattern, using the previously stored lookup table. Interesting animations showing the progress of generation in WFC result from blending the results of the pattern renderer for all patterns that might yet still be placed at a location. Animations of these visualizations over time attracted several technical artists (and the present authors) to learn more about WFC.

Generalizing the role of the pattern classifier, we can imagine other functions which decide how to represent a local patch of pattern placements. Again, we can imagine the use of a deep convolutional neural network to map a small grid of pattern identifier integers into an rich display in the output. Although the pattern renderer's input datatype is fixed, the output can be whatever artist-visible datatype was used as input to the pattern classifier (whether that be image pixels, game object identifiers, or a parameter vector for a downstream content generator).

If additional annotation layers are used in the source images (as in the texture-by-numbers application mentioned above or the player path data present in some VGLC data), it is reasonable to expect that the output of the generator could also have these annotation layers. For platformer level generation, the system could output not only a tile-based map design, but also a representation of which parts of the map the generator expected the player to actually reach.

\subsection{Adjacency Relation}

The basic constraints in WFC are expressed as valid adjacencies between patterns. This is a generalization of the adjacencies between individual tiles: the pattern classifier captures additional adjacency information about the local space, much as an image filter kernel can, and allows the adjacency relation function to infer more general relations between tiles. We prefer that some tiles be allowed to be placed next to each other, such as placing a flower in the middle of a garden. While other adjacencies are non-preferred: the flowers should not be growing out of the middle of a carpeted room.

We can characterize the method used to learn the adjacency legality as Most General Generalization (MGG), the inverse of classic Least General Generalization (LGG) inductive inference technique~\cite{plotkin1971further}. Gumin's implementation simply allows any tile-compatible overlapping patterns to be placed adjacent to one another, even if they were never seen adjacent in the single source image. A side effect of this is that any pattern adjacencies seen in the source image (which are tile-compatible by construction) must be considered valid for the generator to use later.

While MGG might appear as simple parsing and tallying, something too simple to be considered as machine learning, it is useful to compare this approach with classic machine learning techniques like Naive Bayes~\cite[Chap.~20]{aima}. Naive Bayes classifiers are trained with no more sophistication than tallying how often each feature was associated with each class.

The art of constructing the single source image for Gumin's WFC often involves some careful design to include all of the patterns that are preferred and none that are non-preferred. By allowing for multiple positive and negative examples and using a slightly altered learning strategy, we show how this meticulous work can be replaced with a conversation that elaborates on past examples.

\begin{figure}
  \includegraphics[width=3in]{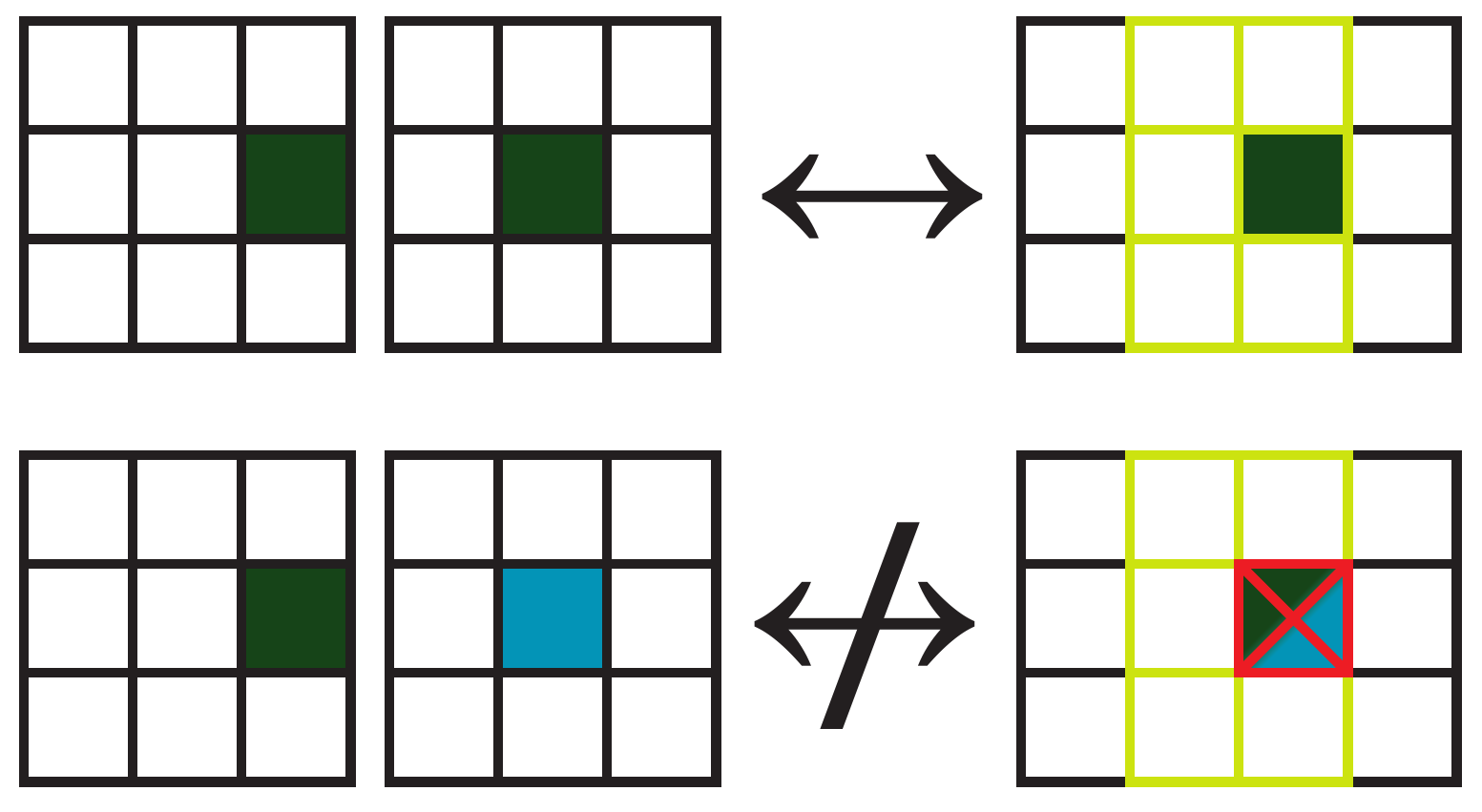}
\caption{An example of a pattern overlap, with a [0,1] offset. The top pair of patterns is a legal overlap, because the intersection of the two patterns matches. The bottom pair of patterns is not a legal overlap, because the blue tile and the green tile conflict. Only the top adjacency is in the legal adjacency set.}
\label{fig:intersections}
\end{figure}

Gumin's pattern classifier function implicitly captures the relationships between patterns in the training data. The first and most absolute distinction is between legal and illegal overlaps: because the patterns in the OverlappingModel need to be able to be placed on top of each other without contradictions, some patterns will never be legal neighbors: if one $3\times3$ pattern has a blue center tile, while another $3\times3$ pattern has a green right tile, the green-right-tile pattern can never be legally placed to the left of the blue-center-tile pattern (Fig.~\ref{fig:intersections}).

The set of all possible patterns that can be assembled from even a small set of tiles can be very large. For example, four distinct tile types can be arranged into a grid $4^9 = 262\,144$ ways. The time and space used by the constraint-based generation approached in WFC scales unfavorably with the number of pattern identifiers, so a much smaller number of patterns is strongly desirable. At the same time, building a lookup table indexed by pairs of all possible patterns is also usually impractical.

Gumin's MGG learning strategy is hardly the only option possible even using the default pattern classifier. A LGG learning strategy would say to only allow those adjacencies explicitly demonstrated in the source image. However, this highly-constrained alternative might not allow any new output to be constructed that was not an exact copy of the source image. Likely, the ideal amount of generalization falls somewhere between these extremes.

\begin{figure}
  \includegraphics[width=3.3in]{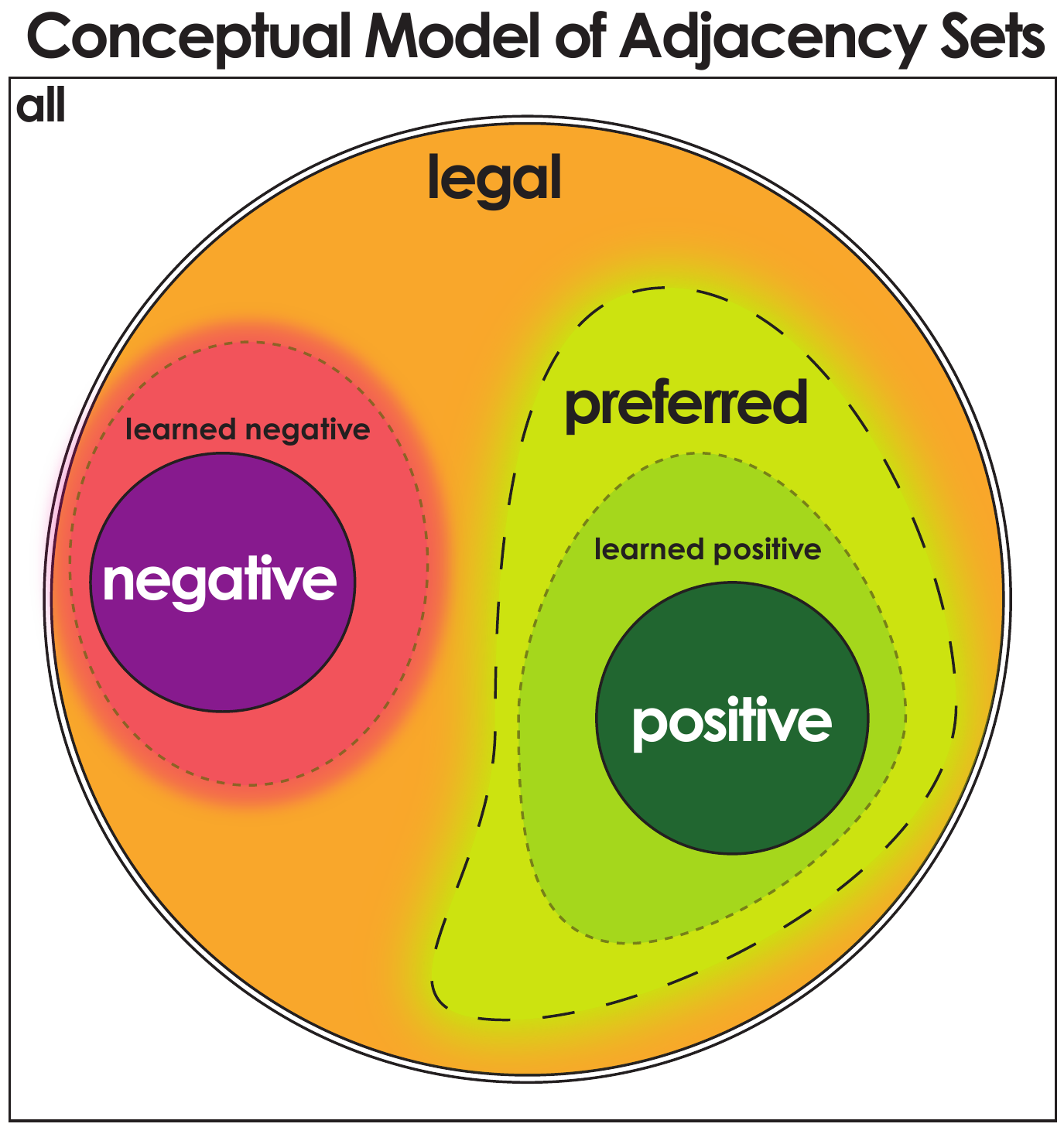}
  \caption{\emph{Adjacency Sets:} The all set contains all of the possible pattern-offset-pattern adjacency triples, using the patterns observed in the image. The legal set contains all of the adjacency triples that can overlap without collision. The positive set contains the adjacencies directly observed in the image. The preferred set is the adjacencies that the artist wants to include. The negative set is the adjacencies that the artist explicitly forbids. The goal is for the valid set (that the machine learns) to match the preferred set (that the artist wants). The original WFC assumes that the set of preferred patterns is identical to the set of legal patterns, and therefore that valid and legal are the same, while we disambiguate them.
Note that this is describing the adjacencies between \emph{observed patterns} (not to be confused with observed adjacencies): the set of all patterns is a superset of the set of observed patterns.
}
\label{fig:adjacency_sets}
\end{figure}

In a discriminative learning setup, we might consider all adjacencies explicitly demonstrated in positive example images to be therefore positive examples for the learned adjacency relation. Likewise, a negative example image needs to demonstrate at least one adjacency that would be considered invalid by the learned relation. We refer to the artist's intended set of allowed relations the preferred set. See Fig.~\ref{fig:adjacency_sets} for an illustration of the relation between the preferred set, the (overlap-compatible) legal set, and the set of all possible pattern pairs to be considered for adjacency.

Gumin's MGG strategy effectively assumes that the preferred set is equal to the legal set, and therefore operates exclusively on the legal set. It does not try to generalize from the observed set or to infer non-observed but possibly still preferred adjacencies. Artists can attempt to adjust the legal set to match the preferred set, but this requires a combination of technical reasoning and trial-and-error iteration. Or, alternately, they could switch to using the SimpleTiled model for which they directly (and with considerable tedium) specify the complete set of allowed adjacencies.

This is a limitation of the learning strategy, not the WFC algorithm itself. The output of the {\tt agrees()} validity function in the original code just checks if two patterns can legally overlap, but any arbitrary adjacency validity function (which accepts two patterns and returns a Boolean) can be substituted here. As long as the validity function can be computed over all pairs of patterns, it can act as the whitelist for the constraint domains, without changing the WFC propagation code itself. The long-range constraint propagation in WFC is more important for its high success rate without backtracking, rather than the heuristic used~\cite{Karth:2017:WCS:3102071.3110566}.

To prototype a variation of WFC supporting an alternate pattern classifier, pattern renderer, and adjacency validity function, we initially developed a surrogate implementation of WFC in the MiniZinc constraint programming language. Later, we integrated the specific ability to generate with a customized pattern adjacency whitelist into a direct Python-language clone of Gumin's original WFC algorithm.

\section{Setting Up Discriminative Learning}
\label{sec:discriminative_learning}

As discussed above, the original approach in WFC is to define the adjacency validity function with the most permissive possible way, using the legal adjacency set as the valid set. Among other drawbacks, this requires careful curation of the patterns so that every adjacency in the legal set is acceptable. While allowing very expressive results from a single source image, there are many preferred sets that are difficult to express in this manner. However, this is just one of the many possible strategies.

An anomaly-detection strategy, such as a one-class Support Vector Machine~\cite{914517}, might allow the set of valid adjacencies to more closely approximate the ideal preferred set, allowing the artist to use patterns with a much larger legal adjacency set.

In this section, we consider the presence of possible negative examples. By removing adjacency pairs that the artist explicitly flagged as undesirable, we can more precisely determine the valid set.

As depicted in Fig.~\ref{fig:adjacency_sets}, the sets of adjacencies allow us to make this distinction clearer: the valid set can be learned from any of these sets. By default, WFC uses the legal adjacency set, but the preferred-valid set can be adjusted to include more or less of the legal set, with corresponding effects for the generation.

\subsection{Machine Learning Setup}

In addition to the single positive source image used by the original WFC, we introduce the possibility of using more source images. Some of these are additional positive examples: it can be easier to express new adjacencies while avoiding unwanted ones by adding a completely separate positive example. In the positive examples, every adjacency is considered to be valid, as usual.

In contrast, sometimes it is easier to specify just the negative adjacencies to be removed from the valid set. In a negative example, at least one adjacency in them is considered to be invalid.

Note that these additional example images do not need to be equal in size. In fact, a tiny negative example showing just the undesirable pairing would work well to let an artist carve out one bad location in an otherwise satisfactory design.

Finally, we have the validity function: a function that takes two patterns (plus how they are related in space, e.g.\ up/down/left/right) and outputs a Boolean evaluation of whether or not their adjacency is valid. In the original WFC, this is simply an overlapping test: given this offset, are there any conflicts in the intersection of the two patterns? However, as suggested above, there are more sophisticated validity functions that also produce viable results. 

\subsection{Human Artist Setup}

In our mixed-initiative training approach, we expect the artist to provide at least one positive example to start the process. The image should demonstrate the local tiles that might be used by the generator, but it does not need to demonstrate all preferred-valid adjacencies. Note that providing one example is the typical workflow for WFC (unmodified, Gumin's code does not accept more than one example). However, instead of expecting the artist to continue to iterate by changing this one example, which can quickly grow complex, the artist can isolate each individual contribution.

Initially, we set the whitelist of valid adjacencies to be fully permissive (MGG), covering the legal adjacencies of the known patterns. From this, we generate a small portfolio of outputs to sample the current design space of the generator. Even a single work sample is often enough to spur the next round of interaction.

The artist reviews the portfolio to find problems. They can add one or more generated outputs to the negative example set directly, crop an example to make a more focused negative example, or hand-create a clarifying example. If additional positive examples are desired to increase variety, those can also be added (although they may immediately prompt the need for negative examples to address over-generation).

With the new batch of source images, we retrain each of the learned functions: the pattern classifier, the adjacency validity function, and the pattern renderer. There are several options for possible ML techniques, which we will explore in more detail below. The update pattern classifier defines the space of patterns that might be placed by the generator. The newly-learned validity function defines the updated whitelist used in the constraints. The updated pattern renderer might even display existing pattern grids in a new way. As before, we sample a portfolio. The artist repeats the process until they are satisfied with the work samples.

The result is a generative system with a design space that has been sculpted to the artist's requirements, all without the artist needing to understand or alter any unfamiliar machine learning algorithms. 

\section{Worked Example}
\label{sec:worked_example}

\begin{figure*}
\includegraphics[width=7in]{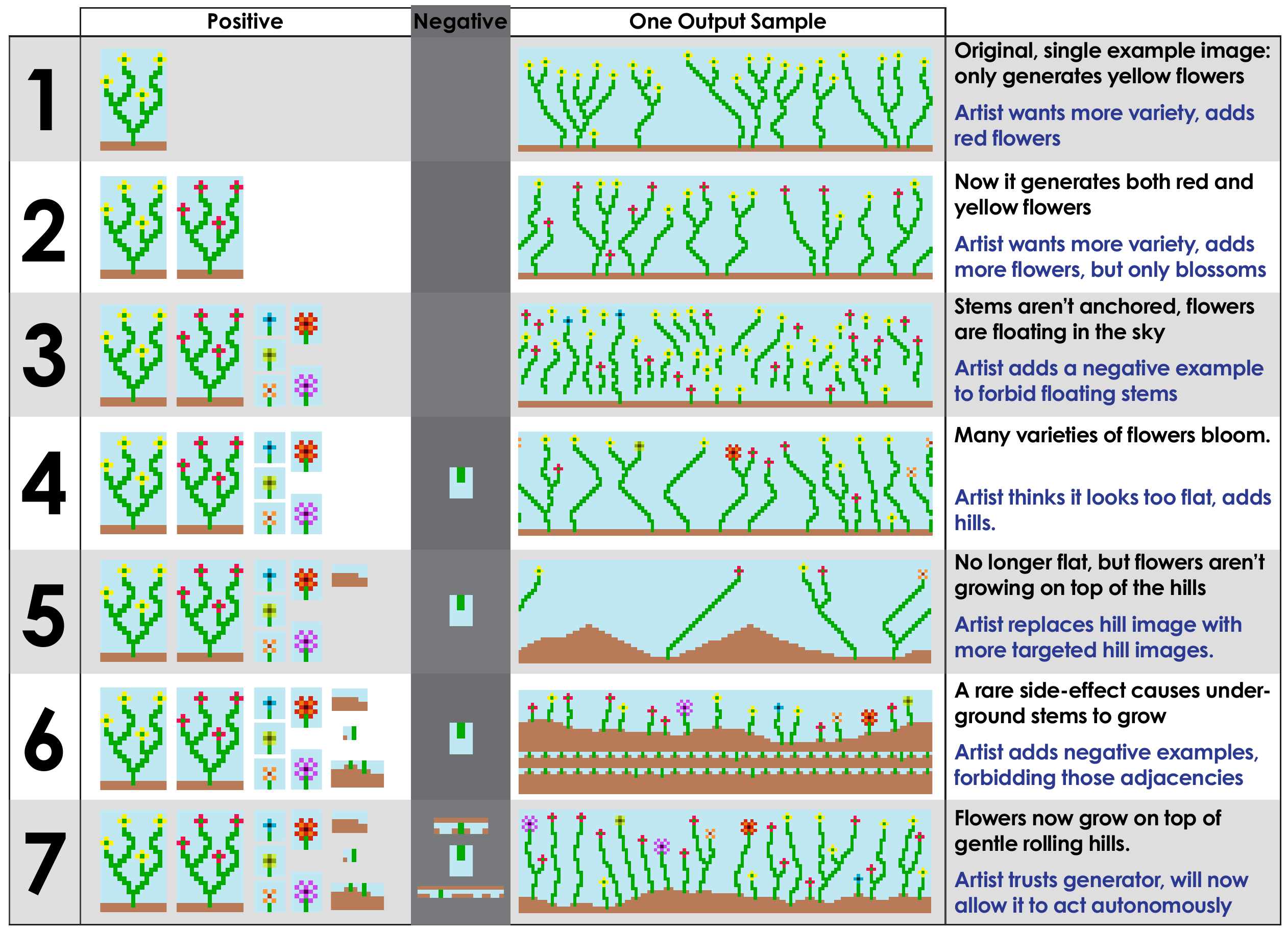}
\caption{A worked example of the mixed-initiative conversational teaching model process. The artist observes the results of each step (top black text) and makes a change for the next step (lower blue text). Each step adds either positive or negative examples. The source images for each output can be seen in the columns on the left, with a representative output image to the right.}
\label{fig:artist_workflow}
\end{figure*}

In this section we walk through an example run of the conversational interaction an artist has with WFC when using a discriminative learning setup. The conversation takes place over several iterations that are visually represented in Fig.~\ref{fig:artist_workflow}. All outputs shown were generated by executing a minimally modified Python clone of Gumin's C\# WFC implementation with an altered pattern catalog (from the pattern classifier) and adjacency whitelist each time. The resulting pattern grids are rendered with the pattern renderer.

In this running example, we make use of a refinement to the MGG strategy used in Gumin's implementation. Rather than simply allowing all patterns which agree on their overlapping tiles, we allow all such patterns \emph{except} those taken from negative examples. Indeed, this is still the most general generalization possible under the extra constraints.

Working through the conversation, our artist begins Iteration 1 with the algorithm by supplying a single positive example. Here we use the \texttt{flowers} example taken from Gumin's public repository. MGG learns the legality relations exactly like the original WFC. The generator is then run to produce a work sample. Observing this, the artist decides that the image needs more colorful flowers.

In Iteration 2, the artist augments the positive image set with a second image, having repainted the flowers to be red. While the original WFC code does not accept more than one input, adding additional patterns only required minimal code changes. In the resulting work sample, now both red and yellow flowers are seen (new patterns were made available to the generator). However, the artist is still interested in seeing more flower variety.

Rather than creating more examples by copy-pasting the original tiles again, this time the artist creates a number of smaller samples that focus exactly on what they want to add to the composition. These extra tiny examples might throw off statistics in a generatively trained model. In the work sample for Iteration 3, the new flowers are present but a surprising new phenomenon arises. The possibility of floating stems results from the particulars of WFC's default pattern classifier and adjacency relation function learning. The artist is not concerned with these implementation details and wishes simply to fix the problem with additional examples.

By selecting and cropping a $3\times4$ region of the last work sample, the artist creates the focused negative example that they use in Iteration 4. MGG, now with the extra constraint from the negative example, no longer considers floating stems to be a possibility despite the fact that this pattern can be identified in the input by the pattern classifier. The work sample is now free from obvious flaws.

Having previously only considered very small training examples, the artist notes a particular feature of the larger generated output. The ground appears uninterestingly flat. In Iteration 5, the artist provides a small positive example of sloped hills, hoping the generator will invent a rolling landscape. However, the work sample for this iteration suggests that the generator has not picked up the generality of the idea from the single tiny example---it only knows how to build continuous ramps without any flowers.

In iteration 6, a few more positive examples show that stems can be placed on hills and that the bumps of hills can be isolated (not always part of larger ramps). However, in rare circumstances the generator will now place stems underground. This would not have been spotted without examining many possible outputs, highlighting the importance of a tool that allows the artist to give feedback on more than one example output.

Finally, in Iteration 7, the artist is able to get the look they preferred. Adding negative examples to take care of the edge cases is easy and can be done without adjusting the earlier source images. Testing shows that the generator is reliably producing usable images. Because of the iterations, the artist now has enough trust in the generator to allow it to perform future generation tasks without supervision. The learned pattern classifier function, pattern renderer function, and adjacency legality function compactly summarize the learning from the interaction with the artist.

In this worked example, every new training example added beyond the first is a direct response to something observed in (or observed to be missing from) concrete images produced by the previous generator. Many demonstrate patterns that are not what the generator should produce in the future, even if it is not realized that this was the case earlier. Instead of iterating to produce a carefully curated set of 100\% valid examples, we make progress by adding focused clarifications.

\section{Conclusion}

The fundamental tension in PCGML is that the effort to craft enough training data for effective machine learning might undermine the motivation to use PCGML in the first place. This makes many machine learning approaches impractical: even when the design goal is flexibility (rather than nominally infinite content) the sheer amount of training data required can be daunting.

However, existing approaches to single-example PCG such as WaveFunctionCollapse suggests that small training data generators are possible. When we combine them with a discriminative learning strategy, we can leverage the usefulness of focused negative examples. As our worked example of the conversational teaching model shows, an artist can intuitively make targeted changes without being overly concerned about maintaining a representative distribution or disturbing earlier, carefully planned patterns just to fix a rare edge case.

Combining PCGML with mixed-initiative design assistance tools can enable artists to sculpt a generator's design space. Rather than building just one high-quality artifact, the artist can train a generator in iterative step to the point where they trust it for unsupervised generation.

\section {Acknowledgements}
The authors wish to thank Adam Summerville for the extensive feedback which greatly improved this research.